\documentclass{svproc}
\usepackage{url}
\usepackage{multicol}

\usepackage[letterpaper]{geometry}
\usepackage{amsmath}
\usepackage{graphicx}
\usepackage{color}
\usepackage{cite}

\usepackage{amsthm}
\usepackage{adjustbox}

\newcommand{\vA}{\mathbf{A}}
\newcommand{\vb}{\mathbf{b}}
\newcommand{\vc}{\mathbf{c}}
\newcommand{\vx}{\mathbf{x}}
\newcommand{\vz}{\mathbf{z}}
\newcommand{\RR}{\mathbf{R}}

\begin{document}
\mainmatter              
\title{\Large (Machine) Learning to Improve the Empirical \\ Performance of Discrete Algorithms\thanks{Supported by NSF grants DMS-1818969 and HDR TRIPODS} }
\titlerunning{Learning to improve algorithms}  
%
\author{Imran Adham\thanks{UC Davis Computer Science.} \and Jes\'us A. De Loera\thanks{UC Davis Mathematics.}  \and
Zhenyang Zhang\thanks{UC Davis Mathematics.}}
\authorrunning{Imran Adham et al.} 
%
%
\institute{UC Davis, Davis CA 95616, USA}

\maketitle              

\begin{abstract}
This paper discusses a  data-driven, empirically-based framework to make algorithmic decisions or recommendations without expert knowledge. We improve  the performance of two algorithmic case studies: the selection of a pivot rule for the Simplex method and the selection  of an all-pair shortest paths algorithm. We train machine learning methods to select the optimal algorithm for given data without human expert opinion. We use two types of techniques, neural networks and boosted decision trees. We concluded, based on our experiments, that:

1) Our selection framework recommends various pivot rules that improve overall total performance over just using a fixed default pivot rule. 
 Over many years experts identified steepest-edge pivot rule as a favorite pivot rule. Our data analysis corroborates that the number of iterations by steepest-edge 
is no more than 4 percent more than the optimal selection which corroborates human expert knowledge, but this time the knowledge was obtained using machine learning. Here our recommendation system is best when using gradient boosted trees.

2) For the all-pairs shortest path problem, the models trained made a large improvement and our selection is on average .07 percent away from the optimal choice. 
The conclusions do not seem to be affected by the machine learning method we used.

We tried to make a parallel analysis of both algorithmic problems, but it is clear that there are intrinsic differences. For example, in the 
all-pairs shortest path problem the graph density is a reasonable predictor, but there is no analogous single parameter for 
decisions in the Simplex method. 
\end{abstract}
\section{Introduction} What is the best way to select an algorithm? Two different algorithms for the same computational task have difference performances: one algorithm is better on some inputs, but worse on the others. Over the years there have been various theoretical frameworks answering this question. \emph{Worst-case analysis} aims to find the extreme instances that strain the performance the most. \emph{Average-case analysis} on the other hand assumes that input instances come from a fixed probability distribution, thus we can talk about average running time or average complexity. More recently, the \emph{Smooth analysis} is a hybrid of the worst-case and average-case analysis of algorithms where one measures the maximum over inputs of the expected performance of an algorithm under small random perturbations of that input. The performance of many algorithms varies dramatically on the types of input one provides, thus the theoretical evaluations often say nothing useful for the non-expert user. {\em How is a non-expert user supposed to make the right algorithmic choices when a large number of choices are possible? 
How can someone make reasonable consistent choices of  parameters for tuning complicated algorithms?}  Here we propose an ML-based methodology to make those choices in the absence of an expert or to corroborate what an expert can suggest.

Consider for example the very famous Simplex method of G. Dantzig \cite{dantzig1998linear}. This is a well-studied algorithm, researchers have found the worst-case behavior of Simplex algorithm is exponential, for most known deterministic pivot rules \cite{Klee1970HOWGI, JEROSLOW1973367, GOLDFARB1979277, inbook,murty1980,Amenta96deformedproducts,10.1007/978-3-642-20807-2_16} and randomized pivot rules \cite{G_rtner_2007, Kalai1997}. On the other hand, under a specific probability distribution for input instances, the average running time of Simplex algorithm is polynomial in terms of the input size \cite{Borgwardt1982}. Similarly, the smooth analysis shows that the Simplex method is efficient \cite{Dadush_2018}. Despite the theoretical success, neither of the three theoretical evaluations matches the \emph{empirical performance} of the Simplex method, which is known to be very fast in practice. Today the Simplex method has been investigated and improved enormously from its original version \cite{bixby2001}.  It is known that the running time or number of iterations for the Simplex method depends not just on the input data, but how we tune the algorithm itself. E.g., what choice of pivot rule shall we make? This is a question that has been answered by experts by fixing a default pivot rule, which often performs well, but may not be always the optimal choice (this choice is often steepest edge pivot rule). As a proof of concept we demonstrate
how machine learning can recover the hard-won wisdom of experts.

The purpose of our paper is to  discuss a pragmatic framework for empirical algorithm selection tuning and comparison. In the present article we demonstrate a \emph{machine learning-based selection and tuning of algorithms}. Our framework is data-driven, empirically-based, and can help non-experts make reasonable consistent algorithmic decisions without prior knowledge of the algorithms. Users of algorithmic methods often have no knowledge of the worst-case examples, nor can they assume to know the exact distribution 
of their data. Users only have access to data sets. The simple principle we propose here is that, if one has sufficiently many data instances, one can create a practical machine learning recommendation system to efficiently automate the selection of algorithms or their parameter configurations for concrete data sets, with the intention to speed up computation. We picked two case studies to illustrate the framework, but it would apply almost in the same way to other algorithms where the input is based on matrices.

Algorithm selection has seen a strong surge in both practical and theoretical research and we only touch the fraction of the literature that we know deals with algorithm similar to our case studies (for much more we recommend \cite{lagoudakis+littman, yang2018oboe,gupta2017pac,balcan, JMLR:v18:16-558} and the many references therein). Several authors have been directly concerned with algorithm selection and tuning for discrete algorithmic problems (see e.g., \cite{khaliletal2017, balcanetal18, andrychowicz2016learning} and the many references therein). The papers \cite{Bengioetal2018,Smith-survey1999} are great 
surveys of uses of learning in combinatorial optimization. In \cite{Bertsimas_2020} the authors redefine mixed integer convex optimization problems as a multi-class classification problem where the machine learning predictor gives insights on the optimal solution. Dai et al. \cite{KhalilDZDS17} develop a method to learn heuristics over graph problems. Several authors have proposed ways to use machine learning to select the best branching rules (see \cite{Alvarez,khaliletal2017}). Machine learning methods have also been useful in aiding the selection of reformulations and decompositions for mixed-integer optimization~\cite{Bonamietal2018,Kruberetal17}. Some libraries organize data for various NP-hard tasks (where the aim is to predict how long an algorithm will take to solve concrete instances of NP-complete problems, or to choose best approximation schemes tailored by instances) \cite{nudelman04,Bischletal16,kotthoffHO17}. In fact the approach we present here is a simplification 
of the empirical hardness model to predict the running time of algorithms applied to improve logic satisfiability (SAT) 
solvers \cite{Leyton-BrownHHX14, Eggenspergeretal2018}. There are also now a number of well-established software implementations for algorithm 
tuning (see \cite{Eggenspergeretal2019,FeurerKES0H19} and the many references therein).

\subsection*{\textbf{Our contributions}}

In this work we present two case studies of ML-algorithm selection, where we see the behavior is clearly dependent on the right choice of algorithm or algorithm version:   

First, the Simplex method. It is widely used in solving linear programming (LP) problems. Geometrically, Simplex algorithm starts on a vertex of the feasible region (which is a polytope), and generates a path via improving edges until optimum is reached. A pivot rule helps to decide which improving edge to pick if there are multiple choices. In this case, we are interested in applying different machine learning models to study and improve the choice among five pivoting rules for the Simplex algorithms on linear programming based on features of different LP instances.

Second, we do algorithm selection on the problem of computing  the shortest paths between all pairs of vertices on a graph. The All-Pairs Shortest Path (APSP) algorithms that we consider for the comparison are All-Pairs Dijkstra \cite{dijkstra1959}; Floyd-Warshall \cite{floyd1962}, which iteratively improves the lengths of shortest paths using dynamic programming; and finally an algorithm proposed by Peng et al. \cite{peng2012}, which is a dynamic programming improvement of All-Pairs Dijkstra to skip extraneous computations.

We demonstrate that the total performance of algorithms, when guided by Machine Learning (ML) decision-making, is clearly faster than using a single static choice for these algorithms. We implemented two ML methodologies, boosted decision trees and neural networks. We tested two different schemes of predicting the fastest algorithm: \emph{direct classification and run time prediction}. In direct classification a machine learning method is trained to predict which algorithm will run the fastest. In the run time prediction setting, a machine learning method is trained to estimate how long an algorithm will run on a particular instance, then we pick the algorithm that is expected to run the quickest. In addition, we tested different data representations and features. We discuss the details in each of the two situations. Next we present the details and in the end we discuss conclusions. 

\section{Case study 1: The Simplex Method}

\subsection{Algorithms}
We begin with some standard definitions related to linear programming and the Simplex algorithm. This introduction is meant to be brief, and we refer to textbooks (see \cite{schrijver1998theory, dantzig1998linear}) for more extensive background knowledge. For the remainder of this section we let $\vA \in \RR^{m \times n}, \vb\in \RR^m, \vc\in \RR^n$ be given.

\begin{definition}
A linear program in standard form is the optimization problem of maximizing $\vc^T\vx$ subject to $\vA \vx = \vb$ and $\vx \geq 0$.
\end{definition}
\begin{definition}
We say that $B\subseteq [n]$ with $|B| = m$ is a \emph{basis} if and only if the columns of $\vA_B$ are linearly independent, or equivalently $\vA_B$ is non-singular. We say $\vx_B$ a \emph{basic feasible solution} with basis $B$ if $\vA\vx_B = \vb, \vx_B \geq 0$ and for all $j \not \in B$: $x_j = 0$.
\end{definition}

\begin{definition}
The vector of \emph{reduced costs} for a basis $B$ is defined as $$\vz^B = \vc - \vA \vA_B^{-1} \vc_B.$$ We say $j \in [n]$ is an \emph{improving pivot} with respect to $B$ if and only if $\vz^B_j > 0$.
\end{definition}

With the definition of an improving pivot, the Simplex method can be summarized as a process of starting with a feasible basis $B$ and updating with improving pivots until no such improving pivots exist. Now we present three basic pivot rules that our experiments consider:
\begin{enumerate}
    \item \emph{Dantzig:} this rule was suggested by Dantzig \cite{dantzig1998linear}. In every iteration Dantzig's rule picks the non-basic variable with the largest positive reduced cost to be the entering variable.
    \item \emph{Greatest Improvement:} this rule picks the improving pivot that results in the largest increment of the objective function.
    \item \emph{Steepest edge:} this rule performs the improving pivot with the largest rate of increment of objective function per distance traveled along the improving edge.
\end{enumerate}

In the following example we briefly explain how different pivot rules choose different pivots using tableaux.
$$
\begin{array}{c}
\text{Variables}\\
z \\ 
w_1 \\
w_2 \\
w_3 
\end{array}\begin{bmatrix}
\begin{array}{c|cccccc|c}
  z & x_1 & x_2 & x_3 & w_1 & w_2 & w_3 & b \\ \hline
  1 & -5 & -4 & -3 & 0 & 0 & 0 &  0 \\ \hline
  0 & 2 & 3 & 1 & 1 & 0 & 0 & 5 \\
  0 & 4 & 1 & 2 & 0 & 1 & 0 & 11 \\
  0 & 3 & 4 & 2 & 0 & 0 & 1 & 8 \\
\end{array}
\end{bmatrix}
$$ 
We can see that $x_1, x_2, x_3$ have negative coefficients and are the potential entering variables. For Dantzig's rule, we pick $x_1$. For greatest improvement, we pick $x_2$ since the increment of objective function by each variable is $x_1 : 12.5, x_2 : \frac{20}{3}, x_3: 12$. And for steepest edge, we pick $x_3$ since the rate of each variable is $x_1 : \frac{5}{\sqrt{30}}, x_2 : \frac{4}{\sqrt{27}}, x_3 : \frac{3}{\sqrt{10}}$.

We study the pivoting strategies for primal Simplex algorithm implemented in DOcplex \cite{docplex}. These include Dantzig's rule, hybrid (DOcplex's default), greatest improvement, steepest edge and devex.  Hybrid is a pivot rule DOcplex implemented as default, which uses Dantzig's rule in the earlier iterations when there are a lot of choices of improving pivots and switch to steepest edge later. Devex is an approximate version of steepest edge developed by P. Harris \cite{devex}. DOcplex also implemented a steepest edge with slack initial norms, which is slightly cheaper in computation. But in our testing, it usually is not better than steepest edge. As a consequence it was not included in the algorithm portfolio.

\subsection{Data Generation}
The existing libraries (MIPLIB 2017 \cite{miplib2017}, NETLIB \cite{netlib} etc.) of linear programming or integer programming are too small for our training purpose. Hence we generated our own data for training and testing. We adapted the algorithms introduced by Bowly et al \cite{Bowly2020}. Their method involves generating constraint matrix $\vA$, and a solution pair $(\alpha, \beta)$. They used $\vA, \alpha, \beta$ to generate the final linear problem maximizing $\vc^T\vx$ subject to $\vA \vx \leq \vb$. For simplicity, we replaced the generation of variable constraint graph by generating Erdős-Rényi (ER) random graphs.

For training and validation set, we generated 24634 instances of linear programming problems with number of constraints ranging from 120 to 200 and number of variables ranging from 50 to 100. For testing, we generate 7279 more instances. Note that these linear programs will most likely be characterized as ``easy" problems by MIPLIB 2017. For the ER random graphs, the parameter $p$ was drawn from $\mathcal{U}\{0.2, 0.8\}$. For other hyperparameters in generating the LP instances, we draw the coefficient mean $\mu_A$ from normal distribution $\mathcal{N}(0, 1)$, coefficient standard deviation $\sigma_A$ from uniform distribution $\mathcal{U}\{1, 10\}$, primal versus slack basis $\gamma$ from $\mathcal{U}\{0.2, 0.8\}$, fractional primal $\lambda$ from $\mathcal{N}(0, 1)$ and Beta fraction $a = 0.5$.

After generating the LP instances, we solve our LP problems using primal Simplex solver in DOcplex with default initialization. We store the number of iterations for each instance using different pivot rules. Note that the LP instances we generate may have degeneracy, and empirically there is a high likelihood of degeneracy where the constraint matrix is low-density.
\subsection{Feature selection}
We have two different ways of choosing features for the linear programming instances. The first method we use is a bag-of-features, where we add features based on heuristics from previous studies on the Simplex method. Apart from $m, n$ the number of constraints and the number of variable, we add three sets of features: variable constraint graph features, coefficient values, and normalized coefficients. Variable constraint graph features include the minimum, maximum, mean, and standard deviation of the degree sequences of variable nodes and constraint nodes. Coefficient values include the statistics of the coefficient matrix $\vA$, the constraint vector $\vb$, and the objective function $\vc$ (i.e. he minimum, maximum, mean, standard deviation, norm of the vector, and the smallest non-zero absolute value). Finally, normalized coefficients are the statistics of row and column normalized coefficients ($\{\frac{\vA_{ij}}{\vb_j}|\vb_j \neq 0\}$ and $\{\frac{\vA_{ij}}{\vc_j} | \vc_j \neq 0\}$) and degree normalized coefficients ($\{\frac{\vb_j}{deg(u_j)}\}$ and $\{\frac{\vc_i}{deg(v_i)}\}$).

The other way we have implemented features related to the coefficient matrix $\vA$, is the Truncated Singular Value Decomposition (SVD), which is a method of dimension reduction \cite{manning08}. The truncated SVD of a matrix $\vA \in \RR^{m\times n}$ returns three matrices $U,\Sigma, V$ such that: 
$$\vA\approx U\Sigma V$$
where $U\in \RR^{m\times k},\Sigma \in \RR^{k\times k}$, and $V \in \RR^{k\times n}$, where $k$ is the number of top singular values to keep. Multiplying $U$ by $\Sigma$ allows for the computation of an $m\times k$ matrix. Applying this procedure again to $(U\Sigma)^\text{T}$ will then compute a $k\times k$ matrix with similar features to the original matrix $\vA$. We choose $k = 20$ in this experiment for the best performance. We still include the features of statistics of the constraint vector $\vb$ and objective function $\vc$.

\subsection{Experiments}
We train four models to choose which pivoting strategies will perform the best on each LP instance. Two models use the bag of features that we choose for LP problems, and the other two use the SVD to replace the features of the coefficient matrix $\vA$. 

\subsubsection{Boosted Trees}

We train two boosted trees to predict the best pivot rule for each LP instance. The first one is an empirical hardness model, that is, for all five pivot rules, we use regression on the features we selected to predict number of iterations that the solver will take using certain pivot rule. The second model is a boosted tree classifier using truncated SVD as features.

\textbf{Bag-of-features boosted trees}
The first model is an empirical hardness model, where we use gradient boosted trees to do regression and predict the number of iterations each pivot rule would cost. Table \ref{xgbr-hyper} shows the hyperparameters for different pivot rules. This model results in a 67.78\% accuracy on the test set with 178.5934 iterations on average.
\begin{table}
\centering
\begin{adjustbox}{center}
\begin{tabular}{|l | c | c | c | c | c|}
\hline 
Hyperparameters & Dantzig & Hybrid  & Devex & Steepest & Greatest \\
\hline \hline
learning rate & 0.1 & 0.1 & 0.1 & 0.1 & 0.05 \\ \hline
\# estimators & 271 & 137 & 173 & 173 & 371 \\ \hline
max depth & 5 &6 &4 & 6 & 6 \\ \hline
min child weight & 6 & 6 & 5 & 4 & 1 \\ \hline
$\gamma$ & 0 & 0 & 0 & 0 & 0.3 \\ \hline
subsample ratio & 1 & 0.8 & 0.8 & 0.9 & 0.8 \\ \hline
column subsample & 1 & 1 & 1 & 0.8 & 0.9 \\ \hline
regularization $\alpha$ & 100 & 10 & 1e-5 & 100 & 1e-5 \\ \hline
\end{tabular}
\end{adjustbox}
\caption{Hyperparameters for each regressor.}
\label{xgbr-hyper}
\end{table}  

\begin{figure}[ht]
 \includegraphics[scale = 0.35]{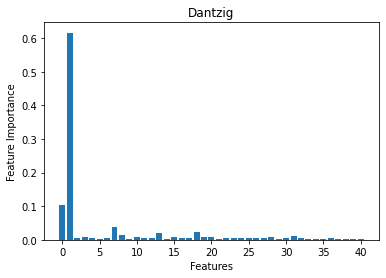}   
 \includegraphics[scale = 0.35]{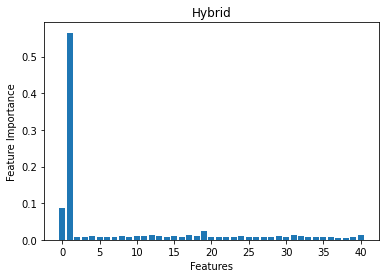}
 \includegraphics[scale = 0.35]{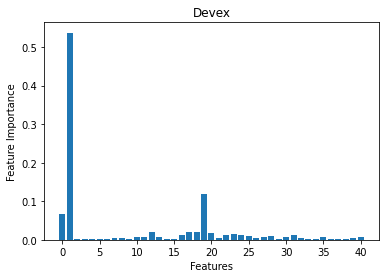}
 \includegraphics[scale = 0.35]{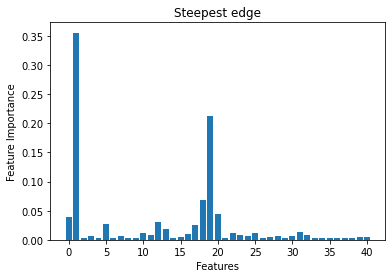}
 \includegraphics[scale = 0.35]{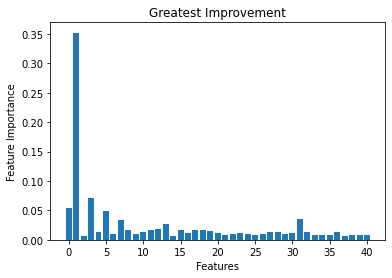}
 \caption{The gain of features for boosted tree regressors.}
 \label{xgbr-features}
\end{figure}

Figure \ref{xgbr-features} shows the gain of features for boosted tree regressors. We can see that apart from number of constraints and number of variables, some of the common features that are important are: maximum number of constraint degree, max and mean of variable degree, min and mean of coefficient matrix $\vA$, min, mean, norm and standard deviation of objective function $\vc$ etc. One could take the subset of important features to train smaller models, which makes the training much faster, but the accuracy will drop to 66.44\% with 178.7804 iterations on average.

\begin{figure}[ht]
\centering
\includegraphics[scale = 0.5]{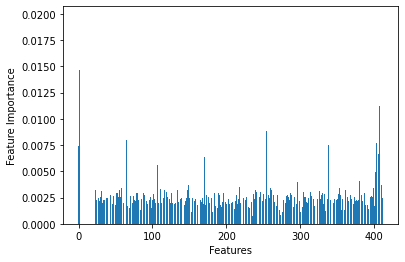}
 \caption{The gain of features for boosted tree classifier.}
\label{xgbc}
\end{figure}

\textbf{Boosted tree classifier} The other boosted tree uses the truncated SVD with $k=20$ as part of the features while keeping the features of constraint vector and objective function. This random forest contains 102 trees with minimum child weight of 5, maximum depth of 5, learning rate of 0.1, subsample and column subsample by tree ratio of 0.8. This model results in a 67.15\% accuracy on the test set with 179.0714 iterations on average. The feature importance is shown in Figure \ref{xgbc}.

As we can see, number of variables and constraints (the first and second feature), as well as features related to constraint vector and objective function are of great importance. Meanwhile, the diagonal entries of the SVD matrix have a relatively high importance. 
\subsubsection{Neural Networks}
We train two models to classify which pivoting strategies will perform the best on each LP instance. The first model uses the bag of features that we choose for LP problems, and the second model uses the truncated SVD matrix to replace the features of the coefficient matrix $\vA$.

\textbf{Bag-of-features Neural Network}
We first train a neural network using features of LP instances we pre-selected. The architecture of the network consists of four hidden layers of ReLU activation function with 64 neurons. Each hidden layer has a dropout of 0.1. The output layer contains five neurons with the softmax activation function. We train the neural network to minimize the categorical cross-entropy loss with the RMSProp optimizer with a learning rate of 0.01 and momentum of 0.2. We train with a batch size of 64 for 50 epochs. This model results in a 62.2\% accuracy on the test set with 179.279 iterations on average. Figure \ref{fig:nn1} (in Appendix) plots the accuracy and loss during each epoch.


\textbf{Truncated SVD Neural Network}
We then train a neural network using truncated SVD matrices as features for coefficient matrix $\vA$ while keeping the features of constraint vector and objective function. The architecture consists of four layers of 512 hidden units with ReLU activation function. The output layer contains 5 neurons with the softmax activation function. We train the neural network to minimize the categorical cross-entropy loss with the ADAM optimizer with a learning rate of 0.001. We train with a batch size of 64 for 100 epochs. This model results in a 72.78\% accuracy on the test set with 179.18 iterations on average. Figure \ref{fig:nn2} (in Appendix) plots the accuracy and loss during each epoch.

\subsection{Comparison of models}

Here we summarize the performance of our models. Table \ref{table:summary} shows the average number of iterations (from the most to the fewest) if we use certain pivot rule or follow our models to solve the LP instances in the test set. It also demonstrates the prediction accuracy of our models. Table \ref{table:compare} shows the instance-wise comparison between our model recommendations with the most popular steepest edge pivot rule. We can see that the best performance of our four models is 69.06\% of the number of iterations steepest edge will take. And they vary on the worst case behavior, with our gradient boosted tree regressor being the most consistent: their worst case will only cost 174.19\% of what steepest edge will perform. We run the Wilcoxon Signed Rank test on each of the test instances between our models and steepest edge pivoting strategy. We can see in Table \ref{table:compare} that except for SVD-20 NN model, the other three models have significant improvement compared to using steepest edge pivoting strategy on all test instances.

\begin{table}[ht]
\centering
\begin{adjustbox}{center}
\begin{tabular}{| l | c | c | c |}
\hline 
 Classifier & Average iterations on test set & Accuracy\\
\hline \hline
Greatest Improvement & 326.1419 & - \\ \hline
Dantzig & 319.7501 & - \\ \hline
Devex & 262.2335 & - \\ \hline
Hybrid & 217.2856 & - \\ \hline
Steepest edge & 179.4161 & - \\ \hline
Bag-of-features NN & 179.279 & 62.2\% \\ \hline
SVD-20 NN & 179.18 & 72.78\% \\ \hline
XGBClassifier & 179.0714 & 70.15\% \\ \hline
XGBRegressor & 178.5934 & 67.78\% \\ \hline
Best in theory & 173.1783 & 100\% \\ \hline
\end{tabular}
\end{adjustbox}
\caption{Summary of average number of iterations and accuracy of each model.}
\label{table:summary}
\end{table}   

\begin{table}[ht]
\centering
\begin{tabular}{| l | c | c | c |}
\hline 
 Classifier & Best & Worst & Wilcoxon test p-value\\
\hline \hline
Bag-of-features NN & 69.06\% & 258.91\%  & $5.37085\times 10^{-13}$\\ \hline
SVD-20 NN & 69.06\% & 210.25\% &  0.91014 \\ \hline
XGBClassifier & 69.06\% & 318.06\% & $6.12158\times 10^{-17}$ \\ \hline
XGBRegressor & 69.06\% & 174.19\% & $1.4\times 10^{-31}$  \\ \hline
\end{tabular}
\caption{Comparison between our models and steepest edge pivot rule on test set per instance.}
\label{table:compare}
\end{table} 

\section{Case study 2: All-Pairs Shortest Path Problem}
\subsection{Algorithms}
The All-Pairs Shortest Path (APSP) algorithms that we consider for the portfolio are:
\begin{enumerate}
    \item All-Pairs Dijkstra's Algorithm \cite{dijkstra1959}: This algorithm simply applies the standard Dijkstra's algorithm for every possible starting node to calculate every possible pair of shortest paths.
    \item Floyd-Warshall \cite{floyd1962}: This algorithm is a dynamic program that stores all the lengths of shortest paths between every pair of nodes. It begins by initializing the path lengths to the weight of the path connecting every node, or $\infty$ if no edge exists. Then it loops through all possible pairs of nodes $n$ times to incrementally update the shortest path between them by checking all intermediary nodes.
    \item Peng \cite{peng2012}: Applies All-Pairs Dijkstra's using dynamic programming to store solutions in order to reduce redundant calculations. In addition, the nodes in the graph are sorted in decreasing order by their degrees in order to maximize the number of calculations that are skipped. The standard Dijkstra's algorithm may be faster if the overhead cost of sorting the nodes is too high on a particular graph.
\end{enumerate}

Two other commonly used APSP algorithms were also considered initially: All-Pairs Bellman-Ford \cite{ford62, bellman58} and Johnson's algorithm \cite{johnson77}. These algorithms have the benefit of being applicable to graphs with negative edge weights, making them more versatile. However, in our testing we noted that these algorithms were always the slowest for all test cases, so these algorithms were not included in the algorithm portfolio.

\subsection{Data Generation}
Graphs for the training and test sets were generated randomly using four methods: Erdős-Rényi (ER) random graphs \cite{ergraph}, Barabási-Albert (BA) random graphs \cite{bagraph}, Watts-Strogatz (WS) small world random graphs \cite{wsgraph}, and finally geometric random graphs. Graphs were generated with nodes between 20 and 1250. After a graph is generated, every edge within it is given a random integer weight between 1 and 100. The parameters for the generation of the graphs were chosen in order to produce a variety in the densities of the graphs while also trying to ensure the graphs are connected. The training set consists of 2309 graphs and the test set contains 1125 graphs with nodes between 20 and 500.

For ER graphs, the parameter $p$ was set to be a random number between $\frac{\ln n}{n}$ and 1. $p=\frac{\ln n}{n}$ is a transition where the graph will likely be connected. For BA graphs, the parameter $m$ is an integer chosen uniformly between 5 and $n-1$. In the WS model, the mean degree $K$ was picked randomly between $\text{ln }n$ and $n-1$, and the parameter $\beta$ was chosen randomly between 0 and 1.

The geometric graphs were constructed by first generating $n$ points within the unit cube. If two points are within a distance of $\epsilon$ of each other, an edge is added between them. The value of $\epsilon$ was chosen to be between $\frac{20}{n}$ and 1.

After generating each graph, we run all three algorithms on them and record how long each algorithm takes to run, along with the algorithm that runs the fastest. In the classification setting, the label for each sample is the algorithm that performed best, and in the regression setting the label is the runtime for the respective algorithm. 10\% of the training data is set aside for validation. Assuming 100\% accuracy in predicting the fastest algorithm, the total runtime of the best algorithm on each graph of the test set is 4356 seconds.

In addition, we test the neural network on a real-world Facebook social network, provided by Stanford \cite{fbgraph}. It contains 4039 nodes and 88,234 edges, and has a topology that is not very well represented by the training set alone.

\begin{figure}
    \centering
    \includegraphics[scale=0.45]{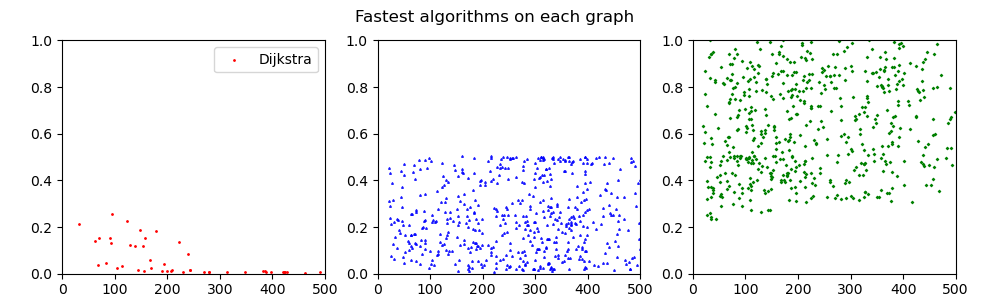}
    \caption{Graphs from the test set plotted as density vs number of nodes and labeled with the fastest algorithm}
    \label{fig:density}
\end{figure}

Figure \ref{fig:density} displays each graph from the test set plotted by its density vs its number of nodes, labeled by the algorithm that runs fastest on it. The figure shows that Dijkstra only runs fastest on graphs with a low number of nodes and low density, Floyd-Warshall tends to run fastest on graphs with high density, and Peng's algorithm is fastest on most lower density graphs. 

\subsection{Feature Selection}
We represent graphs for training the models in two ways. The first method we use is the Truncated Singular Value Decomposition (SVD) of the adjacency matrix as described earlier in the pivot rule selection. The second representation is a sampling of the degree sequence of the graph.

In the Truncated SVD representation, we use $k=20$ as the parameter. We lose information on the number of nodes and edges of the graph, so we add $1/n$ and the density as features. It has been shown that the density of a graph is an important feature to determine when a shortest path algorithm is faster \cite{gallo88}.

Given a degree sequence of a graph, and some parameter $q$, we wish to reduce the degree sequence down to $q$ elements. We take elements with indices $\lfloor \frac{\text{length of degree sequence}}{q} \rfloor i$, for $i=0,1,\ldots ,q-1$. This representation does not maintain the size of the graph and the values are not normalized. So we add $1/n$ as a feature and divide every element of the sequence by $n$. Peng's algorithm is optimized for graphs with few high-degree nodes and many low-degree nodes, so this representation could capture the information necessary to distinguish when an algorithm will be faster.
\subsection{Experiments}
\subsubsection{Boosted Trees}
We train two different boosted trees to predict the fastest APSP algorithm on a given graph. The first model uses the truncated SVD of the adjacency matrix as its features. The second model uses a sample of the degree sequence of the graph. Hyperparameters were tuned via grid search on the parameters of the representation, the maximum depth of the trees, minimum child weight, learning rate, and subsample rate.

\begin{figure}[ht]
    \centering
    \includegraphics[scale=0.4]{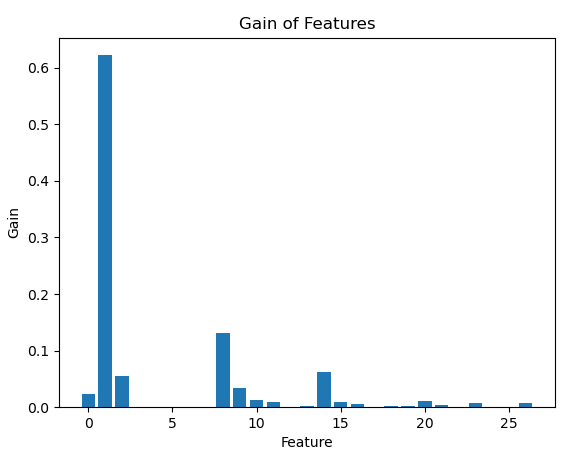}
    \caption{The gain for the SVD boosted tree with $k=5$. First two features are number of nodes followed by graph density. The remaining are elements of truncated SVD matrix.}
    \label{fig:gain_svd_5}
\end{figure}

\textbf{Truncated SVD Boosted Tree}
We begin by training a boosted tree random forest with truncated SVD parameter $k=5$. The forest consists of 64 boosted trees, with maximum depth 6, minimum child weight of 1, and learning rate 0.1. This random forest results in an accuracy of 93.6\% and the total time taken on all the graphs using its predictions is 4359 seconds. Figure \ref{fig:gain_svd_5} plots the gain for each feature. We see that the most important feature is the graph density (second feature), with the diagonal elements of the matrix having relatively high importance.

\begin{figure}[ht]
    \centering
    \includegraphics[scale=0.4]{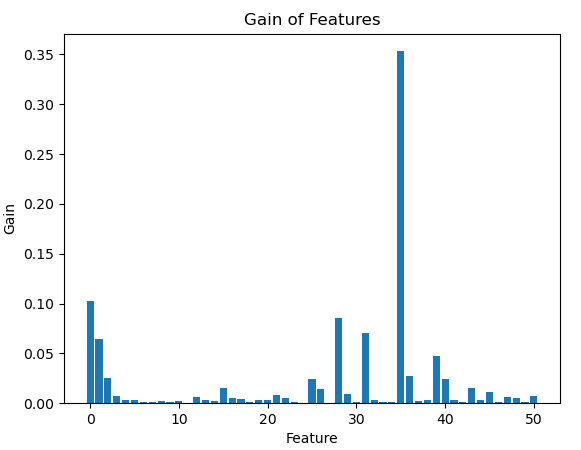}
    \caption{The gain for the degree sequence boosted tree with $q=50$. First feature is the number of nodes, the remaining are the elements of the reduced degree sequence.}
    \label{fig:gain_deg_50}
\end{figure}

\textbf{Degree Sequence Boosted Tree}
Another boosted tree is trained using the sampled degree sequence representation. This random forest contains of 32 trees with a depth of 8, minimum child weight of 1, and learning rate 0.1. The parameter $q$ is set to 50. This forest achieves 93.4\% accuracy with a total time taken of 4359 seconds. The total gain for the features is shown in Figure \ref{fig:gain_deg_50}. The number of nodes (first feature) has high importance, as well as the degrees of the nodes about 3/4 the way through the degree sequence.

\subsubsection{Neural Networks}
In total we train three models to classify which algorithm will perform the best on each instance. The first model is based on a collection of neural networks that predicts the running time of each algorithm. The second uses the truncated SVD as its representation, and the final model uses the degree sequence for its representation.

\textbf{Runtime Prediction Model}
The first model is based on runtime prediction. For each of the three algorithms, a neural network is trained to predict its runtime on a given instance, then the algorithm with the fastest predicted runtime is chosen as the label for the classification. The networks have four hidden layers, the first layer uses the ELU activation function, and the rest use the ReLU activation function. The first layer consists of 512 neurons, and the other three have 256 neurons each. Dropout with $p=0.25$ is added to all the layers. The networks are trained to minimize the mean squared error. The classification accuracy for these networks is 75.4\%. The total runtime of all the graphs in the test set using the algorithms predicted by this classifier is 4594 seconds. Figure \ref{fig:time_pred_history} (in Appendix) plots the accuracy and loss during each epoch.

\textbf{Truncated SVD Neural Network}
The next model trained is a neural network to classify which algorithm will run fastest on each graph using the truncated SVD representation with $k=20$. The architecture of the network consists of five hidden dense layers with 128 neurons each. Dropout with $p=0.5$ is added to each layer to prevent overfitting. The first hidden layer has the ELU activation function, and the other four use the sigmoid activation. The output layer contains three neurons with the softmax activation function. We train the neural network to minimize the categorical cross-entropy loss with the ADAM optimizer with a learning rate of 0.001. We train with a batch size of 64 for 300 epochs. Figure \ref{fig:svd_history} (in Appendix) plots the loss and accuracy of the model at every epoch. This model achieves an accuracy of 93.7\% and using its predictions on the test set causes a total running time of 4360 seconds on the test set.

\textbf{Degree Sequence Neural Network}
The final model we trained is one that uses the sampled degree sequence with $q=50$ as the representation. The architecture for this neural network is two hidden layers with 128 neurons in the first layer, and 64 neurons in the second layer. The first hidden layer uses the exponential linear activation function as before, and the other one uses the sigmoid activation function. Every layer has dropout added with $p=0.5$. The neural network is trained to minimize the categorical cross-entropy loss using the ADAM optimizer with a learning rate of 0.001. The network is trained for 300 epochs with a batch size of 64. The accuracy and loss for this model is plotted in Figure \ref{fig:deg_history} (attached in Appendix). The network has a test accuracy of 93.3\% and the total time taken on the test set is 4364 seconds.\\

\subsection{Comparison of Models}
To compare models and determine which one performs best, we run the Wilcoxon Signed-Rank test on each of the 30 instances of each model. First, comparing the SVD NN model to the Degree Sequence NN model, we find that the degree sequence neural network has a higher accuracy of 93.0\% compared to the SVD's accuracy of 92.1\%. With a $p$ value of $p=0.002$, we conclude that the results are not due to chance. The SVD model takes 4365s to run all test cases while the degree sequence models takes 4366s. Running the Wilcoxon test returns $p=0.2$, and so we determine that they perform similarly even though the degree sequence model has a higher accuracy.\\
We now compare the SVD NN model to the SVD Boosted Tree model. The tree model has an accuracy of 93.1\%. The Wilcoxon test returns $p=0.00002$, so we can conclude that there is a significant difference in the accuracies of these models.

\begin{table}[ht]
\centering
\begin{tabular}{| c | c | c | c | c | c |}
\hline 
-& SVD NN & Deg. Seq. NN & Runtime NN & SVD Tree & Deg. Seq. Tree\\
\hline \hline
SVD NN &- & 0.002 & - & 0.0000197& 0.000008\\ \hline
Deg. Seq. NN & 0.002 & - & - & 0.572& 0.0571\\ \hline
Runtime NN &- & 6053s & - & -&\\ \hline
SVD tree & 0.0000197& 0.572 & 93.1\% &- &0.005\\ \hline
Deg. seq. tree &0.000008 & 0.0571 & 93.3\% & 0.005&-\\ \hline
\end{tabular}
\caption{Summary of p-values for accuracy from Wilcoxon-Signed Rank Test.}
\label{table:acc_p_summary}
\end{table} 

\begin{table}[ht]
\centering
\begin{tabular}{| c | c | c | c | c | c |}
\hline 
-& SVD NN & Deg. Seq. NN & Runtime NN & SVD Tree & Deg. Seq. Tree\\
\hline \hline
SVD NN &- & 0.206 & - & 0.00000173& 0.00000522\\ \hline
Deg. Seq. NN & 0.206 & - & - & 0.00000192& 0.00000388\\ \hline
Runtime NN &- & 6053s & - & -&\\ \hline
SVD tree & 0.00000173& 0.00000192 & 93.1\% &- &0.329\\ \hline
Deg. seq. tree &0.00000522 & 0.00000388 & 93.3\% & 0.329&-\\ \hline
\end{tabular}
\caption{Summary of p-values for total time from Wilcoxon-Signed Rank Test.}
\label{table:time_p_summary}
\end{table} 

\subsubsection{A Real-World Graph}
We tested the SVD classification neural network on the Facebook social network \cite{fbgraph} to verify if the output is correct and to test the generalization of the neural network. When the algorithms are applied to the graph, Dijkstra's algorithm ran in 532s, Peng's algorithm took 40s, and Floyd-Warshall ran in 12,670s. So Peng's algorithm was considerably faster on this graph than the other algorithms. Inputting this graph into the neural network, the outputted probability vector is (0.0000164, 0.999, 0.0000148). The first coordinate represents the probability that Dijkstra's algorithm is fastest, the second coordinate corresponds to Peng's algorithm, and the third coordinate is for Floyd-Warshall. So the neural network is very confident that Peng will run the fastest on the graph, which is supported by the actual runtime of only 40s.

Table \ref{table:apsp_summary} summarizes the results of all the classifiers. We note that all the ML models largely improves on the performance over just using one algorithm, and they perform similarly. Table \ref{table:apsp_variance} summarizes the largest time saved from correct classifications and largest time lost from incorrect classifications for each model by instance. We can see that every model was able to correctly classify the test instance that had the largest impact on overall time saved. Although the neural networks had the highest test accuracy, they were slightly slower compared to the boosted trees due to the fact that they misclassified the more important test cases; boosted trees were able to correctly classify the more important test cases and had overall better performance.

\begin{table}[!htb]
   
    \begin{minipage}{.5\linewidth}
      \centering
        \begin{tabular}{| l | c | c | c |}
\hline 
 Classifier & Total time on test set & Accuracy\\
\hline \hline
Dijkstra & 9165s & - \\ \hline
Peng & 6658s & - \\ \hline
Floyd-Warshall & 6053s & - \\ \hline
SVD tree & 4359s & 93.1\% \\ \hline
Deg. seq. tree & 4359s & 93.3\% \\ \hline
Runtime NN & 4594s & 75.4\% \\ \hline
SVD NN & 4365s & 92.1\% \\ \hline
Deg. seq. NN & 4366s & 93.0\% \\ \hline 
Density heuristic & 4508s & 78.6\% \\ \hline
Best in theory & 4356s & 100\% \\ \hline
\end{tabular}
\caption{Summary of total time and accuracy of each model}
\label{table:apsp_summary}
    \end{minipage}%
    \begin{minipage}{.5\linewidth}
      \centering

        \begin{tabular}{| l | c | c | c |}
\hline 
 Classifier & Largest improvement & Largest deficit\\
\hline \hline
SVD tree & 51.6s & 0.76s \\ \hline
Deg. seq. tree & 51.6s & 0.79s \\ \hline
Runtime NN & 51.6s & 13.7s \\ \hline
SVD NN & 51.6s & 1.8s \\ \hline
Deg. seq. NN & 51.6s & 1.04s \\ \hline 
Density heuristic & 51.6s & 12s \\ \hline
Best in theory & 51.6s & 0s \\ \hline
\end{tabular}
\caption{Largest improvement and largest deficit in time for each model by instance}
\label{table:apsp_variance}
    \end{minipage} 
\end{table}


\section{Conclusion}
In this paper, we show one can rely on ML-methods to predict the performance of different algorithms in different input instances. We can then make recommendations and decide the best algorithm to use in a particular situation.

For the different pivoting strategies for the Simplex algorithm, we find that gradient boosting decision trees work the best in predicting the correct number of iterations. Our  ML-method corroborates what human experts have recovered from their experience that most frequently the steepest-edge pivot rule is a great choice.
Tuning hyperparameters helps to improve the performance of the models, but it is feature engineering that actually improves the models by a huge amount. Throughout the process we learn that certain features, such as variable constraint graph degrees, and coefficients in $A$ and $c$, are more important than other features that people empirically believe (row and column normalized features). Truncated Singular Value Decomposition gives us a convenient way to encode the matrix $A$ into features. This improves the prediction accuracy of our models, but might not necessarily enhance the performance in number of iterations. All of our four models are able to outperform the popular steepest edge pivoting rule by a small edge, and their performance on the test set are pretty close. Although there is a gap between our model and the theoretical optimum, our experiments show that machine learning can help to improve the choice of pivoting strategy. With a proper way to encode linear programs of different dimensions, we might be able to improve the performance of the Simplex algorithm further.

For the problem of computing all-pairs shortest paths we found applying ML techniques to perform algorithm selection vastly improves on the overall performance over selecting an individual algorithm. We found that the method for classification did not greatly affect the performance, neural networks and boosted trees both had very similar performance. We discovered that the density of a graph and the degrees of the nodes are the most important features in selecting algorithms for APSP. Based on Figure \ref{fig:density} it seems as if it might be possible to select Peng's algorithm for graphs with density less than 0.5, and Floyd-Warshall otherwise. But if we use this rule as a classifier, we get a test accuracy of only 78.6\% and a total time taken on the test set of 4508 seconds. This heuristic largely under-performs our ML models, showing that ML can be used to discover deeper, useful patterns in the data to improve results. 

We have presented a very simple machine learning data-driven approach for empirical algorithm selection or parameter tuning that is widely applicable. Given data and a collection of algorithms or parameters from which to choose, our empirical algorithm selection and tuning approach can be utilized to obtain automatic recommendations by almost anyone. We must of course discuss the benefits and shortcomings of our empirical algorithm selection. 

Our approach does not formally prove our selection is optimal for all input instances, but instead only with respect to available data. Although our methods apply to any algorithmic problem with matrices as input we lack a theoretical recipe to choose the features used for training, and will likely change depending on the algorithm. We cannot answer questions such as ``Is there a canonical best ML method?'' or ``What is the optimal neural network architecture (number of layers, activation functions, etc) for a particular algorithm selection problem?''. But this is not a limitation of our approach, rather it is a drawback of the entire theory of machine learning. On the other hand, there are multiple advantages to using our approach. Foremost, it is very basic and simple but improves computation. A user does not require expert-level knowledge of algorithms to make reasonable decisions. Our approach is a pragmatic way to justify algorithmic choices based on available data, and we also provide some consistency and rigor for evaluating algorithms' performance. Moreover, human experts tend to narrow algorithmic choices to one popular default setup which leads to a one-size-fits-all situation.  Our approach allows variability in the choice of algorithm or parameters depending on the concrete instance and, most importantly, results in a clear improvement of running time or computational cost.

\bibliographystyle{plain}
\bibliography{ML.bib}
\section*{Appendix}

\begin{figure}[ht]
\centering
\includegraphics[scale = 0.25]{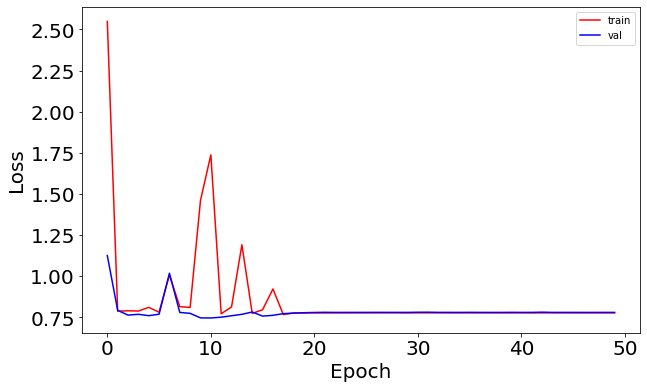}
\includegraphics[scale = 0.25]{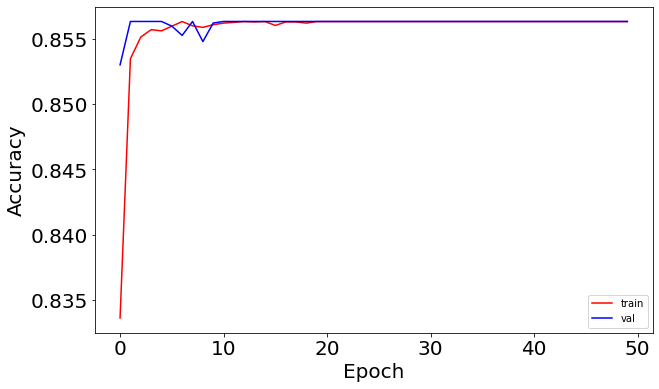}
\caption{The accuracy and loss of the bag of features classification against number of epochs}
\label{fig:nn1}
\end{figure}
\begin{figure}[ht]
\centering
\includegraphics[scale = 0.25]{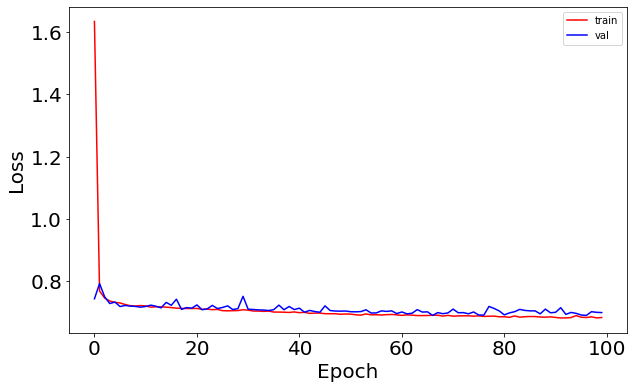}
\includegraphics[scale = 0.25]{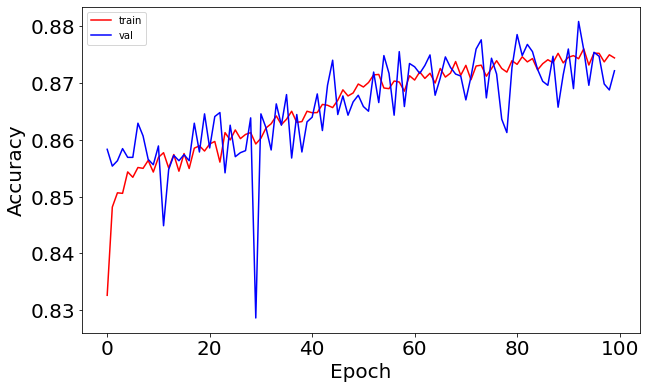}
\caption{The accuracy and loss of the SVD20 classification against number of epochs}
\label{fig:nn2}
\end{figure}

\begin{figure}[ht]
    \centering
    \includegraphics[scale=0.46]{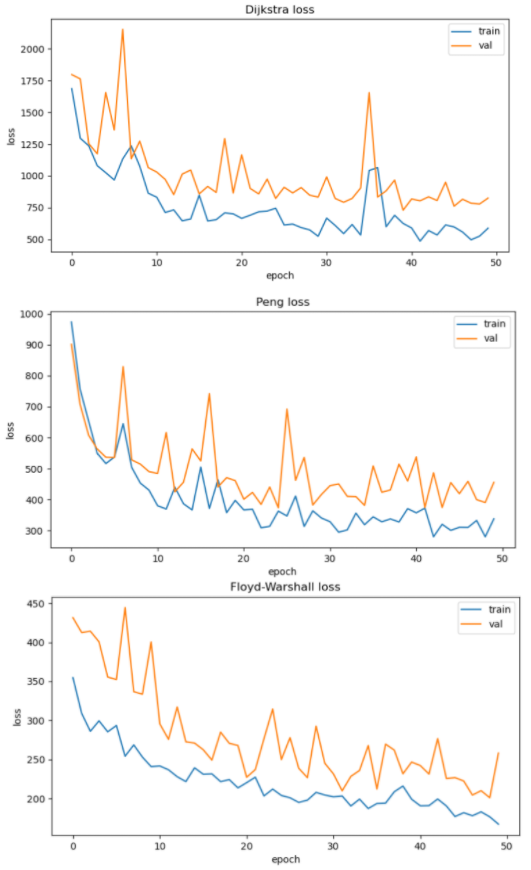}
    \caption{The accuracy and loss of the runtime prediction models at each epoch}
    \label{fig:time_pred_history}
\end{figure}
\begin{figure}[ht]
    \centering
    \includegraphics[scale=0.4]{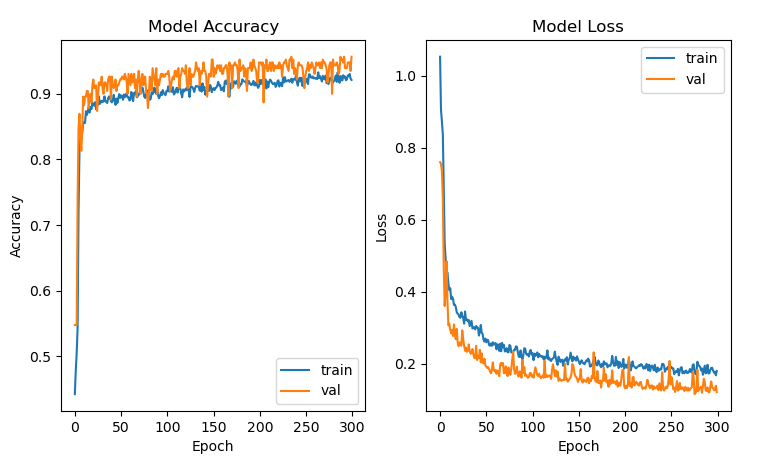}
    \caption{The accuracy and loss of the SVD classification model plotted against the number of epochs during training}
    \label{fig:svd_history}
\end{figure}

\begin{figure}[ht]
    \centering
    \includegraphics[scale=0.4]{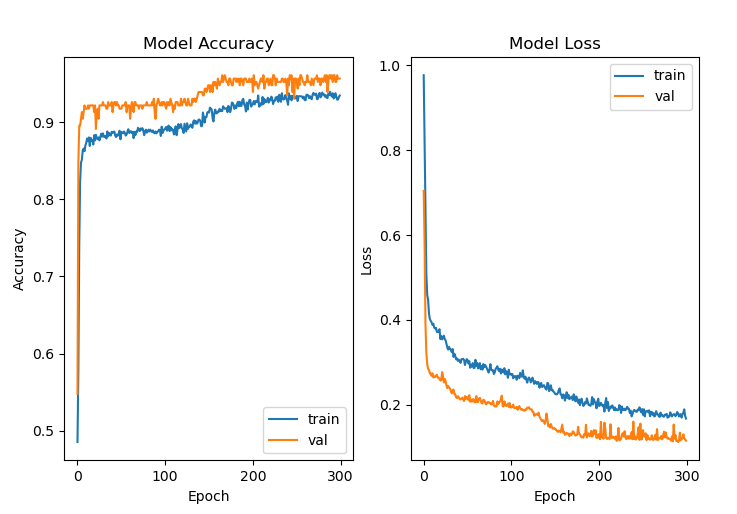}
    \caption{The accuracy and loss of the degree sequence classification model plotted against the number of epochs during training}
    \label{fig:deg_history}
\end{figure}
\end{document}